\documentclass{article}

\usepackage{arxiv}

\usepackage[utf8]{inputenc} % allow utf-8 input
\usepackage[T1]{fontenc}    % use 8-bit T1 fonts
\usepackage{hyperref}       % hyperlinks
\usepackage{url}            % simple URL typesetting
\usepackage{booktabs}       % professional-quality tables
\usepackage{amsfonts}       % blackboard math symbols
\usepackage{nicefrac}       % compact symbols for 1/2, etc.
\usepackage{microtype}      % microtypography
\usepackage{lipsum}
\usepackage{graphicx}
\usepackage{amsmath}
\usepackage{amssymb}
\usepackage[linesnumbered,ruled,lined]{algorithm2e} %http://ctan.org/pkg/algorithm2e
\usepackage{subcaption}
\usepackage{float}

\title{\emph{HSolo}:  Homography from a single affine aware correspondence}

\author{
	Antonio Gonzales \\
	Sandia National Laboratories\\
	Albuquerque, NM \\
	\texttt{aigonza@sandia.gov} \\
		\And
	Cara Monical\\
	Sandia National Laboratories\\
	Albuquerque, NM \\
	\texttt{cmonica@sandia.gov} \\
		\And
	Tony Perkins \\
	Sandia National Laboratories\\
	Albuquerque, NM \\
	\texttt{tperki@sandia.gov} \\
}

\begin{document}
\maketitle

% TODO 
% 1 - update to affine aware, -- CM done
% 2 - sift/surf/ransac rm italics -- CM done
% 3 - check ref 125, dyn update
% 4 - scale and rotation/orientation -- CM done
% 5 - Figure #, capitalized, no parentheses -- CM done
% 6 - eq. (#) -- CM done
% 7 - primes in eq 7 and 8 -- CM done
% 8 - figure readability, font size consistent
% 9 - no parenthesis in det, esp in fig 5 legend -- CM done except fig
% 10 - fig 8 -- dot dash visible in legend
% 11 - NAPSAC is referene is 4 and 19
% 12 - BibTex clean up, consistent abbreviations, author names, etc

\begin{abstract}
The performance of existing robust homography estimation algorithms is highly dependent on the inlier rate of feature point correspondences.  In this paper, we present a novel procedure for homography estimation that is particularly well suited for inlier-poor domains. By utilizing the scale and rotation byproducts created by affine aware feature detectors such as SIFT and SURF, we obtain an initial homography estimate from a single correspondence pair. This estimate allows us to filter the correspondences to an inlier-rich subset for use with a robust estimator.   Especially at low inlier rates, our novel algorithm provides dramatic performance improvements.
\end{abstract}

%I looked at various different papers on homography estimation and they all pretty much start with
%A homography is blah
%It is important for these 15 papers 
%A brief summary of approaches
%These approaches are bad because X
%My new fancy thing magically fixes X
\section{Introduction}
We consider the problem of estimating the homography between a pair of images.  The standard approach produces a list of candidate matches from the features produced from a feature point detector, e.g.~\cite[SIFT]{Lowe:2004:DIF:993451.996342}, and then removes outliers from them.  Outlier removal is most commly done via a robust estimator like random sample consensus~\cite[RANSAC]{ransac81}.  The performance of RANSAC depends heavily on the likelihood of randomly selecting samples of solely inliers.  This approach is infeasible for inlier poor domains.  

We propose the addition of a middle step that leverages the scale and rotation byproducts created by affine aware feature detectors such as SIFT and SURF to remove outliers prior to the RANSAC stage.  The use of these byproducts is both computationally efficient and effective, often allowing RANSAC to remain performant in high outlier cases where it would traditionally be infeasible.

\section{Related research}

Since homography estimation is a fundamental problem of computer vision, numerous approaches and mitigation strategies have been presented.
One such approach is focused on improving the feature descriptor to reduce the number of outliers, e.g.~\cite[ASIFT]{Yu09}.  Alternatively, one can look to improve robust estimators, e.g.~\cite[NAPSAC]{napsac},~\cite[PROSAC]{prosac},~\cite[GC-RANSAC]{gcransac}, and~\cite[MAGSAC++, P-NAPSAC]{magsacpp}.   Recently, Barath \textit{et al.}~\cite[2SIFT]{Barath2019} developed a robust estimator using SIFT correspondences that only requires two correspondences, instead of the usual four.

Furthermore, in some specialized applications, one can exploit geometric or spatial constraints to simplify the model.  For example, camera calibration problems, e.g.~\cite{Barath2019, Kukelova19, Kukelova20,pritts20}, are successfully solved by leveraging constraints of a more restrictive model.  In~\cite{Scaramuzza11}, their objective is to estimate the relative motion of a vehicle from a sequence of images of a single fixed camera. The assumption that a camera is on a vehicle, allows them to use a more restrictive motion model which can be parameterized with a single point correspondence.  The authors of~\cite{Jang20} introduce a spatial clustering technique as an intermediate outlier reduction stage.

It's important to note that our line of research is orthogonal to these others discussed.  It is quite possible to combine our outlier removal process with contemporary feature descriptors, RANSAC variants, and restricted models.  

Object recognition is another fundamental problem of computer vision that often relies on feature point matching~\cite{Ferrari,object1,object2}.  In~\cite{fred}, Rothganger \textit{et al.} develop a framework for 3D object recognition that enforces geometric and appearance constraints between image patches, found by affine aware feature detectors.  These geometric constraints implied the ability to create an affine homography from a single match, which we will use in our method.

\section{Background}
\begin{figure}[!hbtp]
\begin{center}
   \includegraphics[width=3in] 
                   {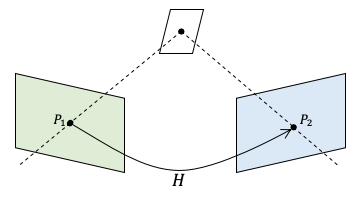}
\end{center}
\caption{The homography \(H\) relates the location, \(P_1\), of a feature in the left image to its location, \(P_2\), in the right.}\label{fig:homography}
\end{figure}

Given a pair of images of the same scene taken from different perspectives, as shown in Figure~\ref{fig:homography}, we're interested in finding the projective transformation between them, which is called a homography.  Robust homography estimation is crucial in many computer vision applications such as image registration~\cite{registration}, and autonomous navigation~\cite{slam}.  

The homography, \(H\), is represented as a \(3\times3\) matrix that transforms \(P_1=[u_1, v_1, 1]^T\) in the first image to its location \(P_2={[u_2, v_2, 1]}^T\) via the transformation
\begin{equation}
\label{eq:H}
\begin{bmatrix}
h_1 & h_2 & h_3\\
h_4 & h_5 & h_6\\
h_7 & h_8 & h_9\\
\end{bmatrix}
\begin{bmatrix}
u_1\\ v_1 \\ 1 \\
\end{bmatrix}
=
\begin{bmatrix}
u_2\\ v_2 \\ 1 \\
\end{bmatrix}.
\end{equation}

The common approach to solving for \(H\) is to identify pairs of identical real world points in the two images, referred to as \textit{correspondences} and use them as constraints on the entries of \(H\).  When inserted into eq.~\eqref{eq:H}, a single correspondence will yield 2 homogeneous linear constraints on \(H\):
\begin{equation}
\label{eq:constraints1}
\begin{split}
u_1h_1 + v_1h_2 + h_3 - u_1u_2h_7 - v_1u_2h_8 - u_2h_9 &= 0 \\
u_1h_4 + v_1h_5 + h_6 - u_1v_2h_7 - v_1v_2h_8 - v_2h_9 &= 0.
\end{split}
\end{equation} 
Given \(4\) correspondences, combined with \(h_9=1.0\) to ensure a unique and non-zero solution, least-squares is used to solve the resulting linear system.

Typically, correspondences are found by using a feature point matching algorithm such as SIFT~\cite{Lowe:2004:DIF:993451.996342} or SURF~\cite{Bay2008346}.  These algorithms produce a set of \textit{candidate} correspondences, only some of which match real world features.  Correspondences that match (resp. don't match) real world features are the \textit{inliers} (resp. \textit{outliers}), and the percentage of true correspondences among the candidates is the \textit{inlier rate}, \(w\).

\(H\) is estimated from the set of candidate correspondences by using a robust estimator, commonly RANSAC.  The RANSAC family varies widely in approach~\cite{choiransac}, but all are an iterative process consisting of:
\begin{enumerate}
\item solving for \(H\) by randomly sampling a minimal set of \(n\) candidate correspondences to do so  and
\item identifying the \textit{support} of \(H\), that is the number of candidate correspondences that \(H\) projects to their expected location, within a small threshold \(\epsilon\).
\end{enumerate}

This process is repeated \(k\) iterations and the homography with the largest support is returned.   Theoretically, this process will succeed if a set of \(n\) inliers is sampled together, and thus 
\begin{equation}
\label{eq:k}
k = \frac{\log(1-p)}{\log(1-w^n)}\
\end{equation}
is chosen to ensure this with probability \(p\).  As seen in eq.~\eqref{eq:k}, the choice of \(k\) is heavily dependent on the inlier rate \(w\), which is generally not known ahead of time.   The value of \(w\) is estimated in advance or dynamically updated at runtime~\cite{prosac}. % TODO - check the dynamically updated reference.

\begin{figure}[H]
\begin{center}
   \includegraphics[width=2in]
                   {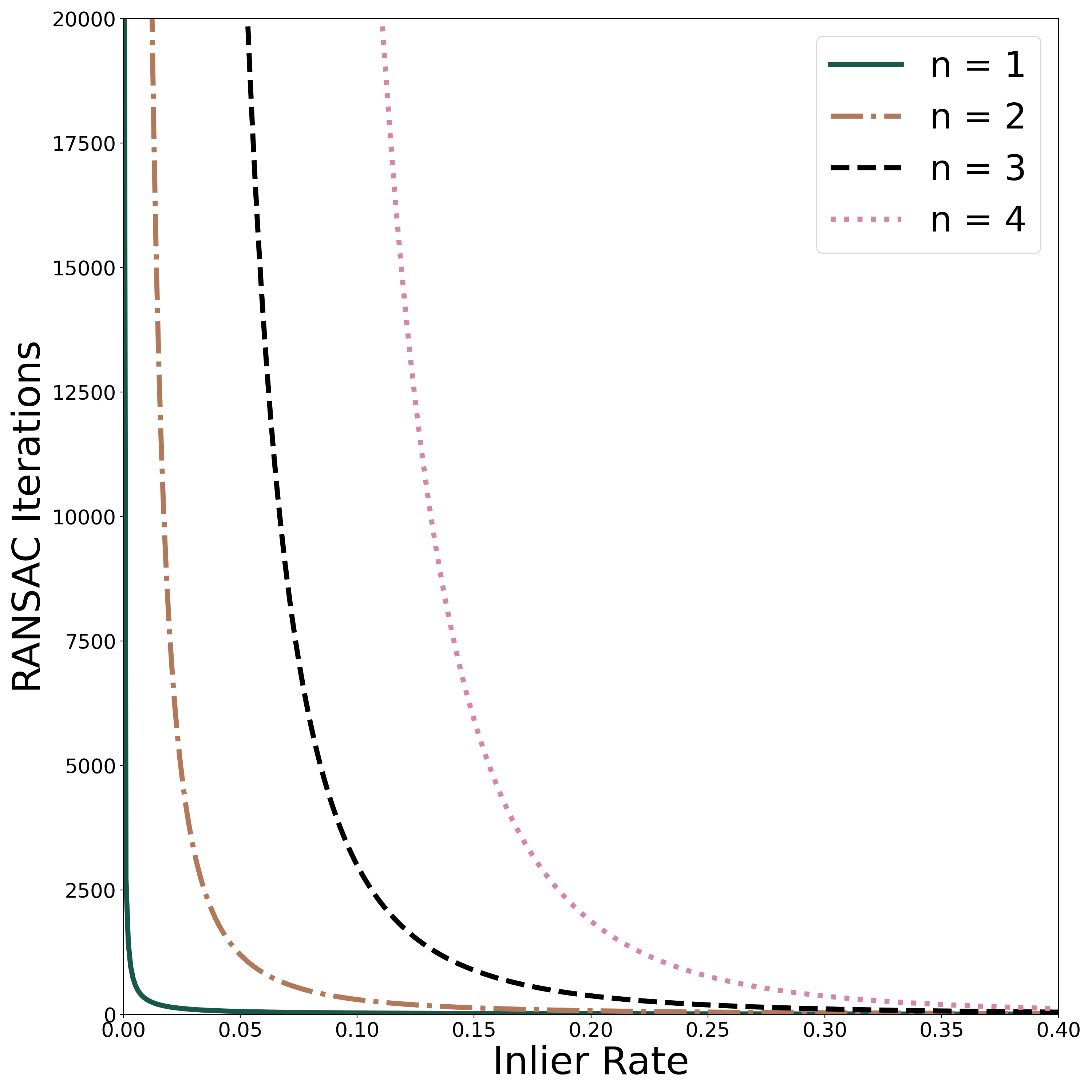}
\end{center}
\caption{RANSAC iterations needed for a \(95\%\) probability of success as \(w\) increases for varying values of \(n\).}
\label{fig:ransack}
\end{figure}

As shown in Figure~\ref{fig:ransack}, RANSAC is very efficient when \(w\) is high, but becomes exponentially expensive as \(w\) decreases. For example, the case of \(n=4\) and \(w=0.03\) requires \(3.7e6\) iterations to achieve a \(95\%\) probability of success.  

\section{\textit{HSolo}}

Our proposed method, \textit{HSolo}, estimates \(H\) from a single correspondence of affine aware features.  We start by assuming \(w = \frac{1}{n}\) and then repeat the following steps until either all correspondences have been chosen or we have performed $k$ iterations: 
\begin{enumerate}
\item randomly choose one correspondence from the candidate set,
\item estimate \(H'\) from the correspondence,
\item use \(H'\) to filter the initial set of correspondences to an inlier-rich subset,
\item use the inlier-rich subset with a robust estimator to calculate \(H\) and its support,
\item update the estimate of \(w\) based on the support of  \(H\) and recalculate \(k\).
\end{enumerate} 

\subsection{Estimate of H from a Single Affine Aware Correspondence}
\label{sec:initialH}

\begin{figure}[H]
\begin{center}
   \includegraphics[width=2.0in]
                   {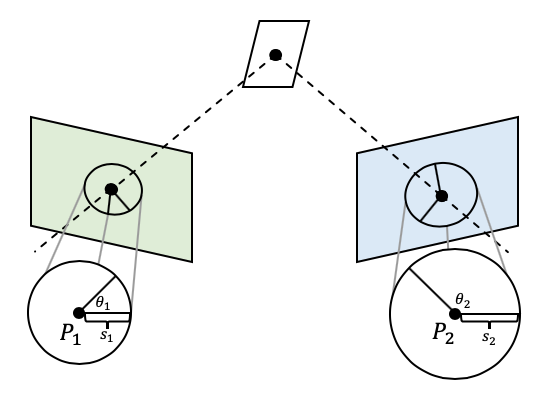}
\end{center}
\caption{Visualization of SIFT features, an example of an affine aware detector.  The values \(\theta_1\) and \(\theta_2\) and \(s_1\) and \(s_2\) are the angle of rotation and feature scale calculated by the SIFT process for points \(P_1\) and \(P_2\) respectively.}\label{fig:correspondence}
\end{figure}

An affine transformation  $A$ may be decomposed as
\begin{equation}
\label{eq:affine}
A=
\begin{bmatrix}
s_x && 0 && 0\\
0 && s_y && 0\\
0 && 0 && 1
\end{bmatrix}
\begin{bmatrix}
\cos\theta && -\sin\theta &&0\\
\sin\theta && \cos\theta && 0 \\
0 && 0 && 1
\end{bmatrix}
\begin{bmatrix}
1 && sk_x && 0\\
sk_y && 1 && 0\\
0 && 0 && 1
\end{bmatrix}
\begin{bmatrix}
1 && 0 && t_x\\
0 && 1 && t_y\\
0 && 0 && 1
\end{bmatrix}
\end{equation}

where \(s_x\) and \(s_y\) are the scaling factors in the x and y directions, \(\theta\) is the angle of rotation, \(sk_x\) and \(sk_y\) are the skew factors, and \(t_x\) and \(t_y\) are the translations.
We say that a feature detector is \textit{affine aware} if the byproducts of detection include scale and rotation about the feature point, as shown in Figure~\ref{fig:correspondence}.   For example, SIFT feature points are defined by a point location \(P = (u,v)\), the angle of rotation \(\theta\), and a scale \(s\).

Rothganger \textit{et al.}~\cite{fred} described a method to calculate the affine homography between a pair of images patches defined by a single correspondence.  First, affine homographies, denoted \(H_1\) and \(H_2\) are created for each point relative to the origin in their respective images.  Then the affine homography, \(H_{12}\), between the image patches is calculated as

\begin{equation}
\label{eq:AHaffine}
H_{12} =
H_1H_2^{-1}.
\end{equation}

It is impossible to compute the exact affine homography between two image patches from the byproducts of affine aware feature detectors.  These byproducts are merely estimates and any error is likely to significantly impact the resulting homography.  Furthermore, to our knowledge, no affine aware feature detectors provide information on the skew factors.  
Despite this, our method leverages the approximation of \(H_{12}\), \(H'\), using transforms \(H_1'\) and \(H_2'\) built directly from the scale and rotation byproducts provided by the affine aware feature detector.  We assume  \(sk_x = sk_y = 0\) and they are left out of eqs. \eqref{eq:Paffine} and \eqref{eq:Qaffine} for simplicity. 

\begin{equation}
\label{eq:Paffine}
H_1' =
\begin{bmatrix}
s_1 & 0 & 0\\
0 & s_1 & 0\\
0 & 0 & 1
\end{bmatrix}
\begin{bmatrix}
\cos\theta_1 & -\sin\theta_1 & 0\\
\sin\theta_1 & \cos\theta_1 &  0\\
0 & 0 & 1
\end{bmatrix}
\begin{bmatrix}
1 && 0 && u_1\\
0 && 1 && v_1\\
0 && 0 && 1
\end{bmatrix}
\end{equation}

\begin{equation}
\label{eq:Qaffine}
H_2' =
\begin{bmatrix}
s_2 & 0 &0\\
0 & s_2 & 0\\
0 & 0 & 1
\end{bmatrix}
\begin{bmatrix}
\cos\theta_2 & -\sin\theta_2 &0\\
\sin\theta_2 & \cos\theta_2 &0\\
0 & 0 & 1
\end{bmatrix}
\begin{bmatrix}
1 && 0 && u_2\\
0 && 1 && v_2\\
0 && 0 && 1
\end{bmatrix}.
\end{equation}

\begin{equation}
\label{eq:AHaffine}
H' =
H_1'H_2'^{-1}.
\end{equation}

\subsection{Filtering Inliers Using \(H'\)}

While our affine approximation \(H'\) is unlikely to be an accurate representation of the projective homography \(H\), we hypothesize that \(H'\) is relatively accurate between the local areas around the points in a correspondence.  Previous work, such as the NAPSAC method of Myatt \textit{et al.}~\cite{napsac}, has shown value in exploiting the assumption that inliers tend to be clustered spatially.  This suggests that if \(H'\) is indeed a good estimate in the local area then we can identify additional inliers that are spatially close to the correspondence used to solve for \(H'\).

\begin{figure}[H]
     \centering
     \begin{subfigure}{0.3\textwidth}
         \centering
         \includegraphics[width=.8\linewidth]{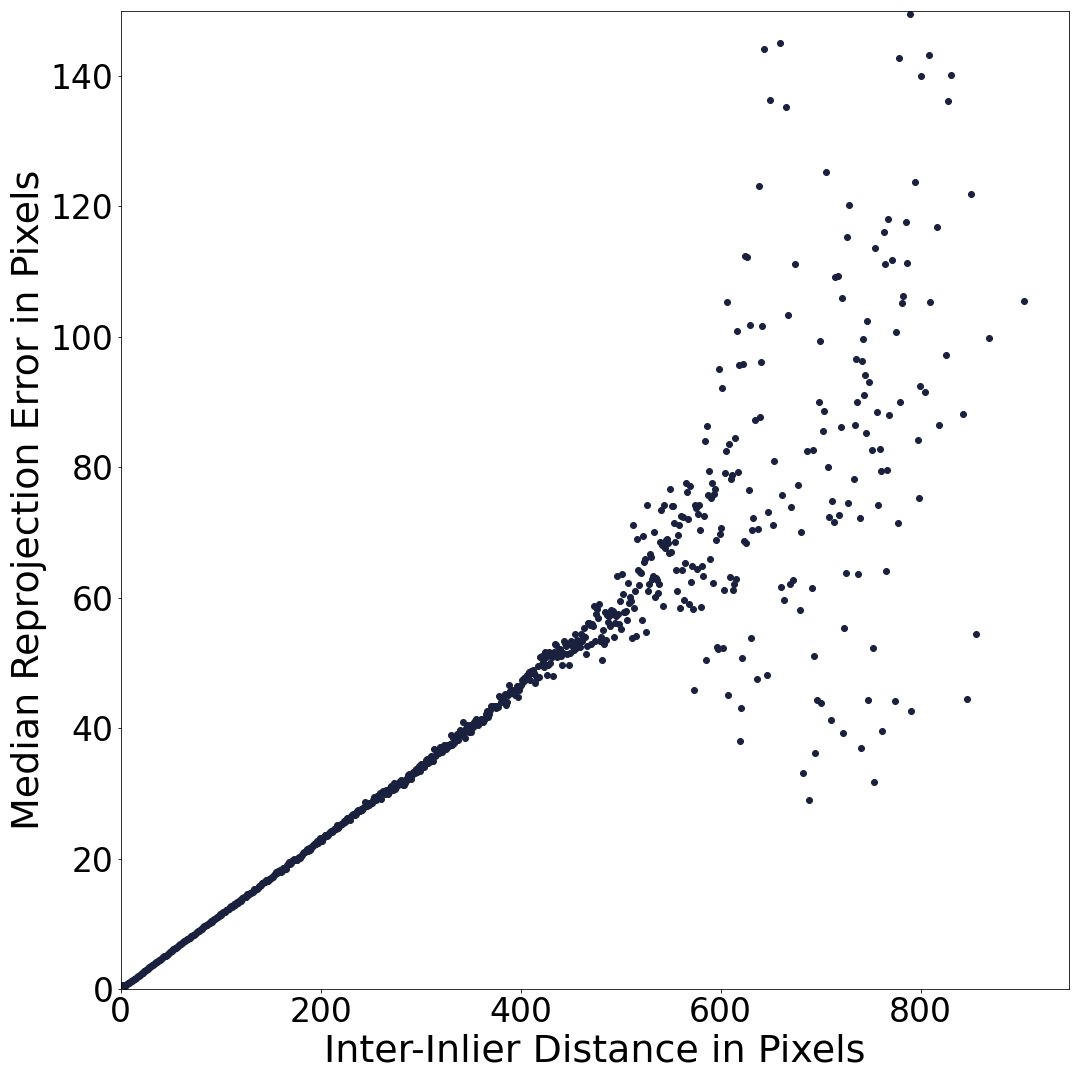}
         \caption{}
     \end{subfigure}
     \begin{subfigure}{0.3\textwidth}
        \centering
         \includegraphics[width=.8\linewidth]{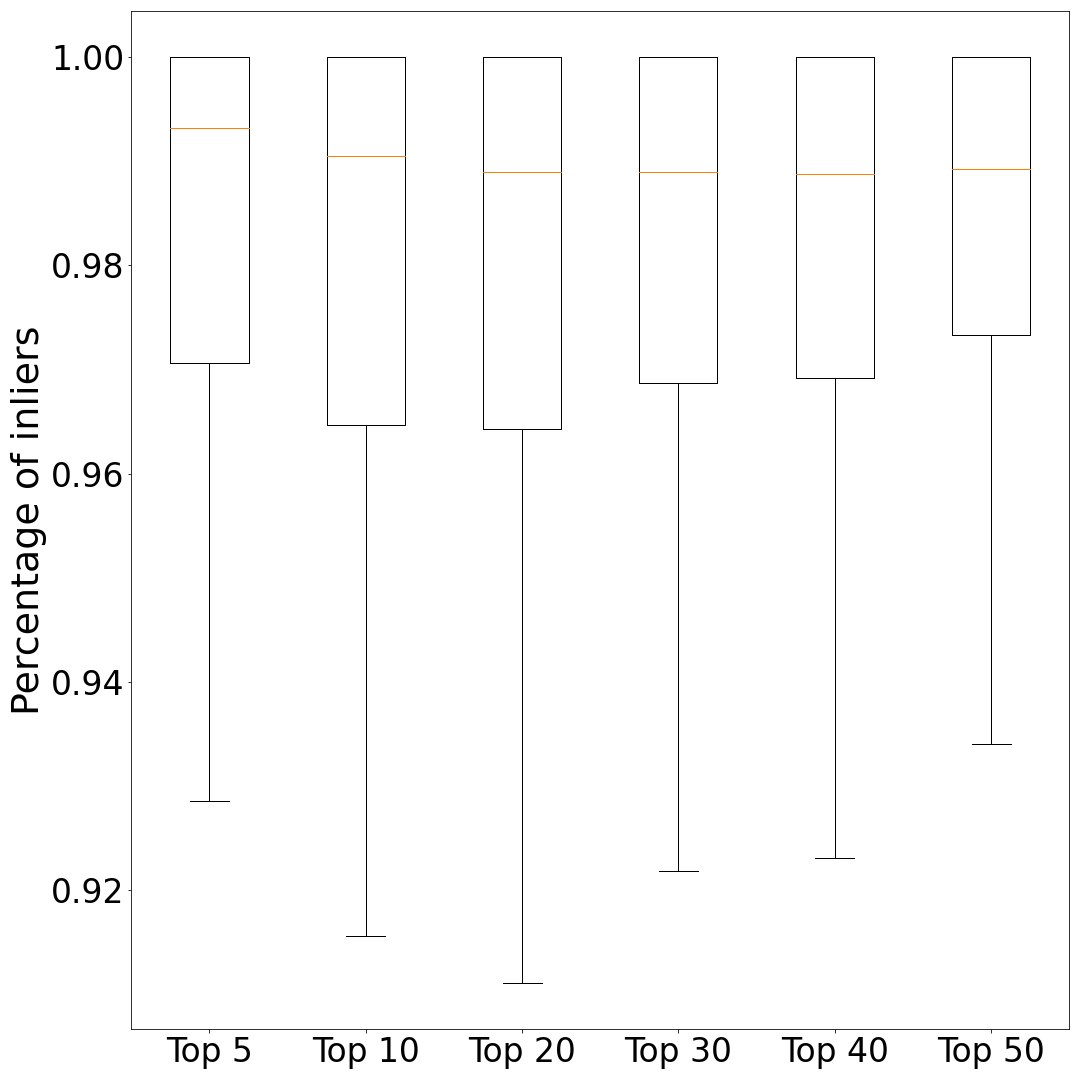}
         \caption{}
     \end{subfigure}
          \begin{subfigure}{0.3\textwidth}
        \centering
         \includegraphics[width=.8\linewidth]{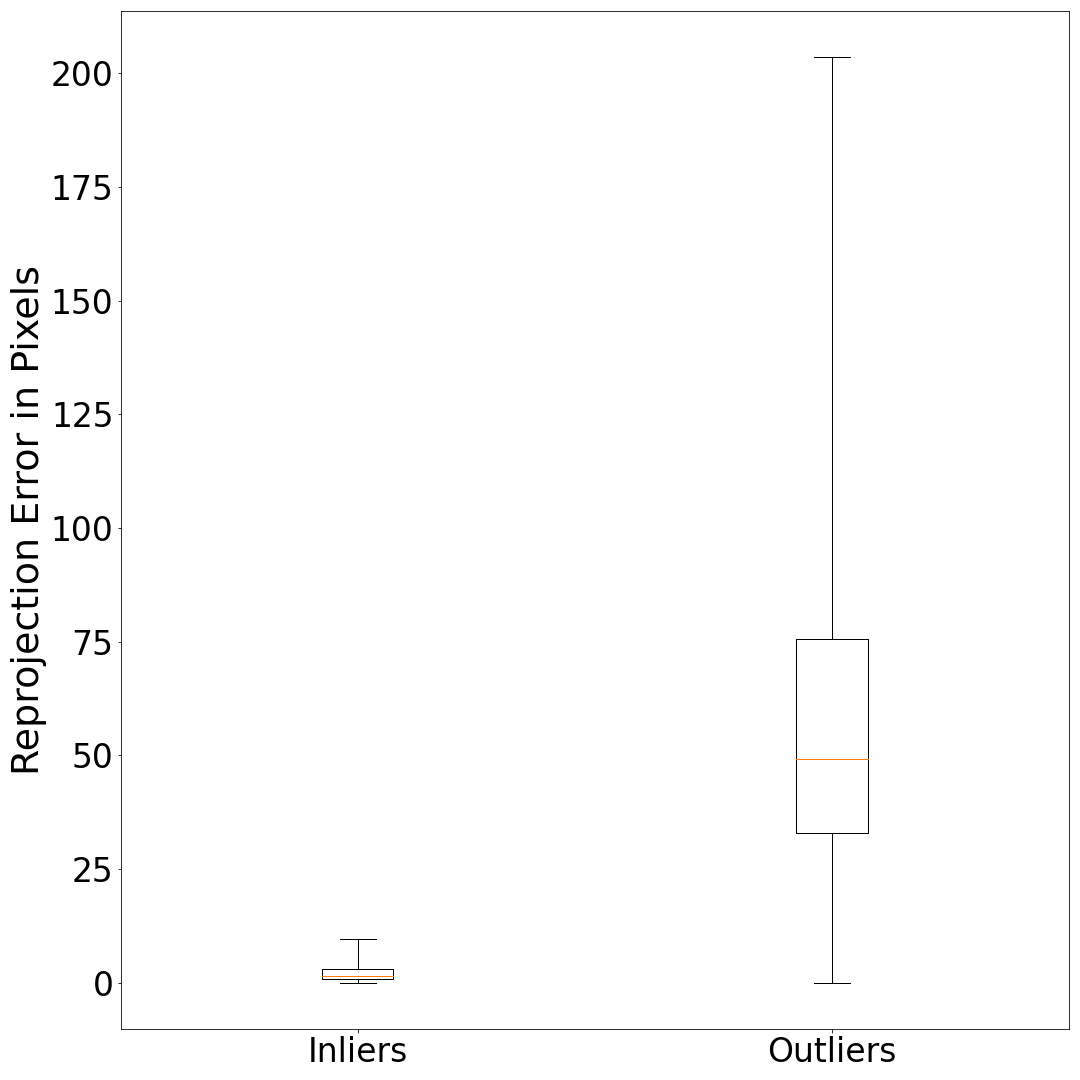}
         \caption{}
     \end{subfigure}
        \caption{(a)  The spatial distance between inliers correlates strongly with the reprojection error for \(H'\), particularly at smaller distances.  (b) Distribution of inlier rate among the top n candidate correspondences by reprojection error for \(H'\).  (c)  The distribution of the median reprojection errors of inliers vs. outliers in the top 20 candidates correspondences.}
        \label{fig:inlier_filtering}
\end{figure}

To test our hypothesis, we examine the inliers in the AdelaideRMF data described in Section~\ref{sub:dataset}.   First, we examine how well \(H'\) estimates \(H\) by taking every known inlier correspondence in the data, solving for \(H'\), and then calculating the reprojection error of all other inlier correspondences.  As shown in Figure~\ref{fig:inlier_filtering}(a), spatially close inliers have lower mean reprojection errors, though the errors are still much too large to accurately estimate \(H\).  

Next we examine if these low projection error inliers are separable among the reprojections errors of the entire candidate correspondence set.  We repeat the same process as above except this time we calculate the reprojection error of all candidate correspondences and sort them in ascending order.  Figure~\ref{fig:inlier_filtering}(b) shows that the large majority of candidate correspondences with the lowest reprojection errors are inliers.  

Additionally, as shown in Figure~\ref{fig:inlier_filtering}(c), there is a large difference in the distribution of reprojection errors in the correspondences with the lowest reprojection errors between inliers and outliers in the candidate correspondence set.  This allows us to set a threshold, \(\epsilon_R\) that determines if a particular set of correspondences is worth further investigation.  The value of \(\epsilon_R\) is an estimate of the upper bound for outliers, \(Q_3+3R\), where $Q_3$ is the 75th percentile and $R$ is the interquartile range for the reprojection errors of the specific data set.

\begin{figure}[H]
	\centering
	\begin{subfigure}{0.3\textwidth}
		\includegraphics[width=2.1in]{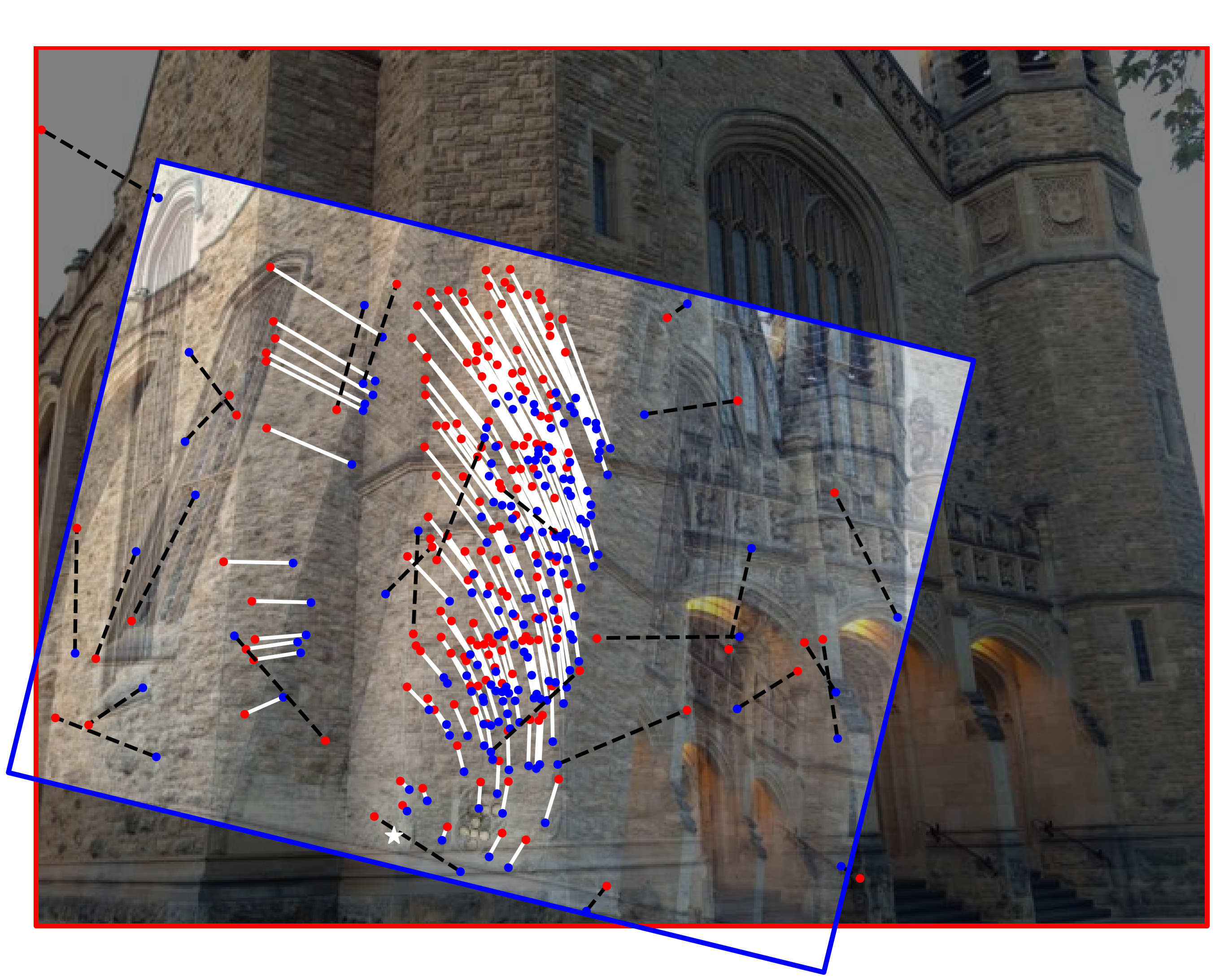}
		 \caption{}
	\end{subfigure}
	\begin{subfigure}{0.3\textwidth}
	\includegraphics[width=2.1in]{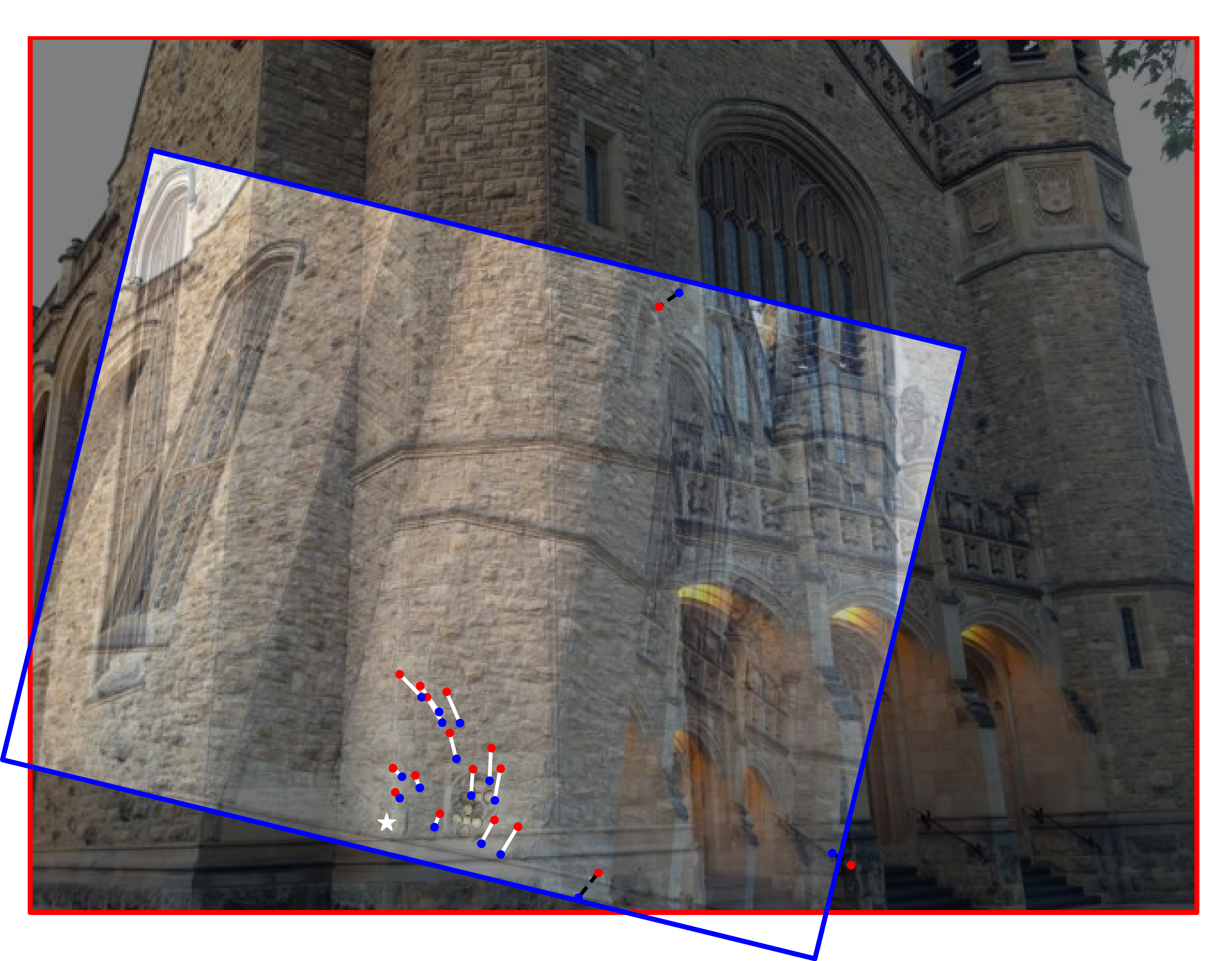}
	 \caption{}
\end{subfigure}
	\begin{subfigure}{0.3\textwidth}
		\includegraphics[width=2.1in]{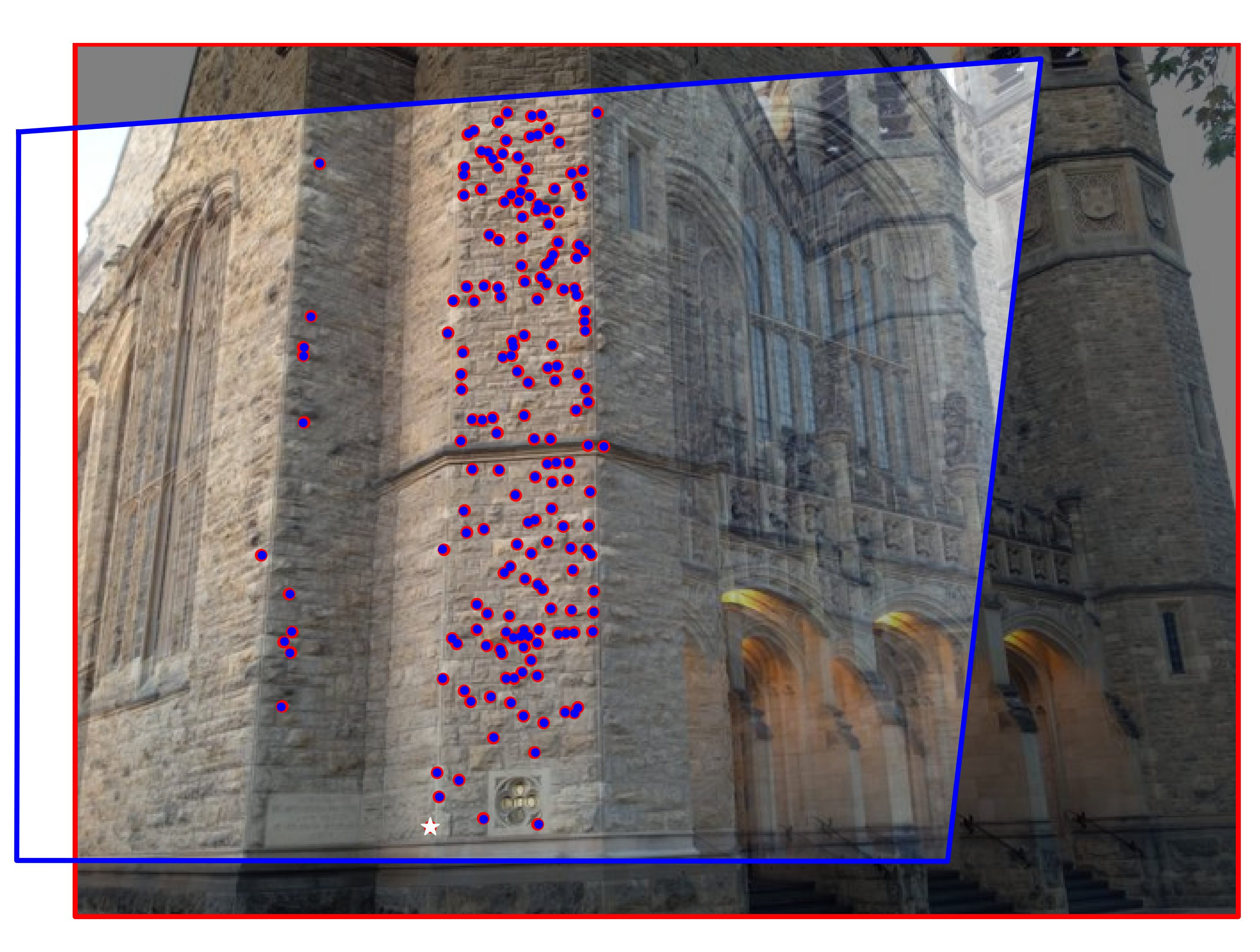}
		 \caption{}
	\end{subfigure}
	\caption{(a) Top 200 correspondences by reprojection error and overlayed images for \(H'\). The star is the single correspondence used to estimate \(H'\). Inliers are shown with solid lines and outliers with the dashed.  (b) Filtered correspondence set of size 20.  (c) Final  correspondence set for \(H\) produced by \textit{HSolo}.}         
	\label{fig:hsolo_image_results}
\end{figure}
These results confirm that the sorted order of the reprojection errors of \(H'\) serve as a filter to find other inliers.  We refer to this set of inlier-rich correspondences as the \textit{filtered correspondence set},  its size as \(n_f\), and its inlier rate as \(w_f\).

%TODO can we remake this figure with |\det H'| = 1.0 in the legend.  Also it feels like the plot segment should be square
%TODO can we have the star be on both endpoints on the correspondence?  Maybe make it blue so it stands out a little more when small?

\subsection{Estimating \(H\) from the Filtered Correspondence Set} \label{sec:estimateH}

The final step of each iteration of \textit{HSolo} is to find the full, projective homography \(H\) using the filtered correspondence set by applying a robust estimator.  In our case, we use the standard 4 correspondence RANSAC method.  We know from the previous section that \(w_f\) for the filtered correspondence set will be high which leads to an efficient estimation of \(H\). 
It is important to note that the number of inlier correspondences in the candidate correspondence set, \(c\), must hold to the relationship \(\tfrac{c}{n_f}\geq w_f\).   If not, RANSAC will not run enough iterations to achieve the desired probability of success.  

 RANSAC draws its samples from the \(n_f\) members of the filtered correspondence set while calculating support against the entire candidate correspondence set.  The number of RANSAC iterations is determined by eq.~\eqref{eq:k} based on \(w_f\) and \(p\).  To increase the numerical stability of RANSAC, we normalize the points as described in~\cite{Hartley97a}.  

At the end of each \textit{HSolo} iteration, we update \(w\) and \(k\) based on the best support of \(H\) found thus far.  After all \textit{HSolo} iterations are complete, we apply an optimizer to the homography to minimize the reprojection errors of the discovered inliers.  
Figure~\ref{fig:hsolo_image_results} shows an example of the filtered correspondence set and final result of \textit{HSolo} on a pair of images from the AdelaideRMF data.

\subsection{Pseudocode}\label{sec:algorithms}

\SetAlFnt{\small\sffamily}
\begin{algorithm}[H]
\label{algo:hsolo}
\SetKwRepeat{Do}{do}{while}%
\KwIn{Correspondence Set \(C\), Filtered Correspondence Set Size \(n_f\), Filtered Inlier Rate \(w_f\),  Error Threshold \(\epsilon\), and Run Inner RANSAC Threshold \(\epsilon_R\)}
\KwOut{Homography \(H\)}

\(w=\frac{1.0}{\mathrm{size}(C)}\)\\
\(k=\) calculate using eq.~\eqref{eq:k}\\
\(\mathrm{bestSupport} = []\)\\
\(H = \textsf{null}\)\\
\(\mathrm{iterNum}=0\)\\
Randomize the order of \(C\)\\
\While{\(\mathrm{iterNum}<\min(k,\mathrm{size}(C))\)}{
\(H'=\) solve using \(C[\mathrm{iterNum}]\) as described in Section \ref{sec:initialH} \\
\(\mathrm{filteredCorrSet}=\) find \(n_f\) lowest error correspondences in \(C\) for \(H'\)\\
\If{\text{median error of } \(\mathrm{filteredCorrSet}\leq\epsilon_R\)} {
\(H'= \mathrm{RANSAC4pt}(\mathrm{filteredCorrSet}, C, w_f, p)\) sample from \(\mathrm{filteredCorrSet}\) but calculate support in \(C\) \\
\(\mathrm{support} = \) members of \(C\) that project within \(\epsilon\) of expected position via \(H'\) \\
\If{\(\mathrm{size}(\mathrm{support}) >\mathrm{size}(\mathrm{bestSupport})\)} {
\(\mathrm{bestSupport}=\mathrm{support}\) \\
\(H = H'\)\\
\(w=\frac{\mathrm{size}(\mathrm{bestSupport})}{\mathrm{size}(C)}\)\\
\(k=\) update using eq.~\eqref{eq:k}
}
}
\(\mathrm{iterNum}\texttt{++}\)\\
}
Apply an optimizer to \(H\).\\
\KwRet{H}

\caption{\textit{HSolo}}
\end{algorithm}

\section{Performance Evaluation}
 \label{sub:dataset}

We evaluate the performance of our proposed method using the AdelaideRMF data set~\cite{wongiccv2011}.  The data consists of 22 image pairs containing a total of 78 homographies, where each homography is defined by a set of manually-identified correspondences.

Our method exploits the scale and rotation features provided by affine aware feature detectors; however, AdelaideRMF only provides the point locations of each correspondence.  In order to create usable ground truth for each image pair, we first find the homography \(H\) via least squares on all the correspondences provided by AdelaideRMF.  Then, we use SIFT to extract feature points and generate correspondences.  We transform all the SIFT features using \(H\) and identify those with a reprojection error of \(<2.0\) pixels.  These correspondences become the ground truth inliers for our experiments.  When multiple homographies are present in an image pair, we evaluate a single homography at a time.  When evaluating a specific homography, inliers from other homographies are set to random locations within the image. 

Table~\ref{tab:table} contains the parameterization used in these experiments.   
Recall from Section~\ref{sec:estimateH}, that the number of true inliers in the correspondence set, \(c\) must satisfy \(\tfrac{c}{n_f} \geq w_f\). Thus, we skip homographies where \(\tfrac{c}{n_f}<0.7\) allowing us to evaluate 73 out of 77.

\begin{table}[H]
	\caption{Parameters used for performance evaluation}
	\centering
	\begin{tabular}{c|ccccc}
		\toprule
		Parameter & $\epsilon$ & $p$ & $w_f$ & $n_f$ & $\epsilon_R$ \\
		\midrule
		Value & 4.0 & 0.95 & 0.7 & 21 & 20 \\
		\bottomrule
	\end{tabular}
	\label{tab:table}
\end{table}

\subsection{\textit{HSolo} Complexity}

The complexity of \textit{HSolo} is directly related to the complexity of the RANSAC.  RANSAC's complexity is dominated by the number of iterations run, \(k_R = \frac{\log(1-p)}{\log(1-w^4)}\) (see eq.~\eqref{eq:k}), and the \(O(n)\) time required to calculate the support for the candidate homography in each iteration.  In the worst case, complexity of RANSAC is 
\begin{equation}
\label{eq:RANSACComplexity}
O(k_Rn).
\end{equation}

\textit{HSolo} runs \(k_H= \tfrac{\log(1-p)}{\log(1-w)}\) iterations each of which requires the search for an \(n_f\) sized filtered correspondence set and \(k_{Hr}=\tfrac{\log(1-p)}{\log(1-w_f^4)}\) RANSAC iterations.  Finding the filtered correspondence set can be done via a partitioning algorithm and a priority queue with complexity \(O(n\log(n_f))\), and thus the complexity of \textit{HSolo} is
\begin{equation}
\label{eq:HSoloComplexity}
O(k_Hn(\log(n_f)+ k_{Hr})).
\end{equation}
\textit{HSolo} will be faster than 4 correspondence RANSAC when
\begin{equation}
%\begin{align*}
\label{eq:HSoloComplexity2}
k_H(\log(n_f)+ k_{Hr}) \lesssim k_R  \Longrightarrow \\
\log(n_f) + \frac{\log(1-p)}{\log(1-w_f^4)} \lesssim \frac{\log(1-w)}{\log(1-w^n)}.
%\end{align*}
\end{equation}

\begin{figure}[H]
   \centering
   \includegraphics[width=2.25in]{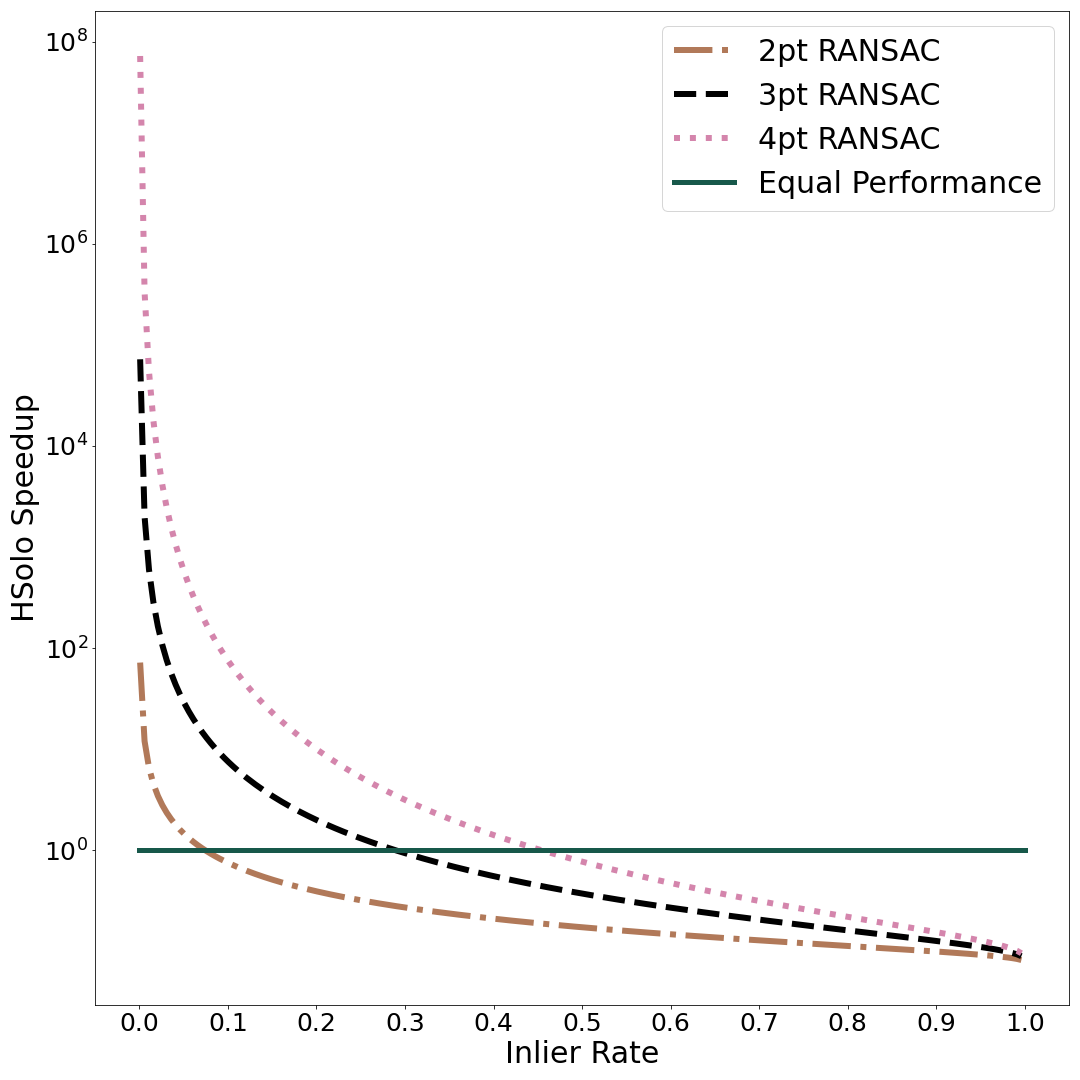}
   \caption{\textit{HSolo} speedup with \(w_f=0.7\) and \(n_f=21\) compared against RANSAC for varying \(n\) as \(w\) increases.  The probability of success is set to \(95\%\).  The horizontal line represents equivalent performance.}\label{fig:hsolospeedup}

\end{figure}

\begin{figure}[H]
	\centering
		\includegraphics[width=4in]{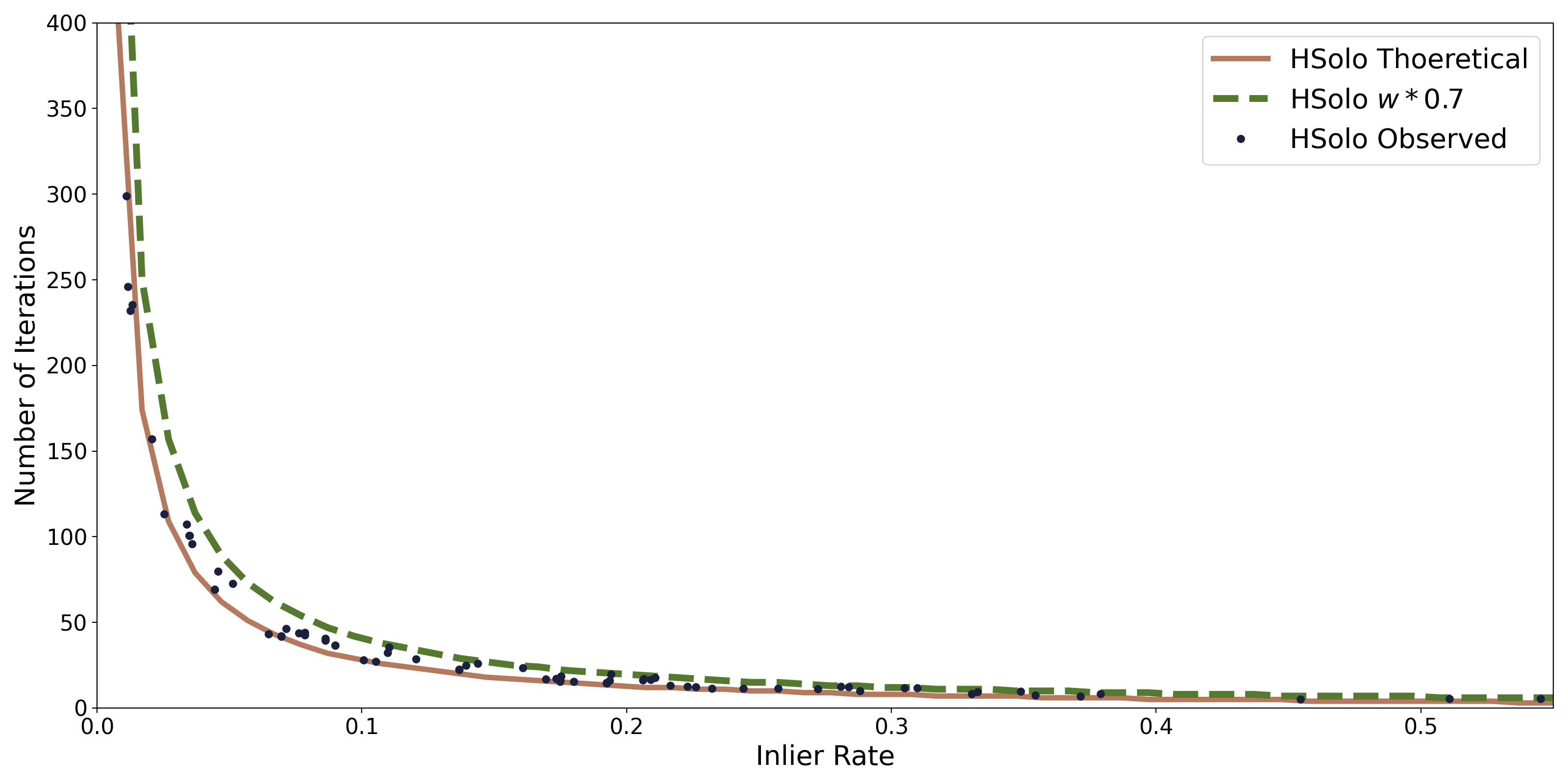}
	\caption{Theoretical number of \textit{HSolo} iterations vs. the observed number required to obtain a correct solution.}
	\label{fig:hsolo_theory_reality}
\end{figure}

As shown in Figure~\ref{fig:hsolospeedup}, \textit{HSolo} provides a significant theoretical speedup over standard RANSAC  at lower values of \(w\).   To evaluate how well our implementation provides these theoretical speedups, we compared the theoretical and observed number of iterations required to obtain a correct solution, as seen in Figure~\ref{fig:hsolo_theory_reality}.

In the ideal case, every inlier in the candidate correspondence set will produce a good solution.  Due to the inherent imprecision in the detection of feature point scale and rotation, it is likely that some inliers will produce an unusable estimate of \(H'\).  Inliers that fail to produce a good solution have the effect of reducing \(w\) which increases the number of iterations required by \textit{HSolo}.  To quantify the impact, we  run 500 trials for each of the homographies with the parameterization listed in Table~\ref{tab:table} and allow \textit{HSolo} to run until it finds a correct solution.   We find that by applying the scaling to the inlier rate, \(w*0.7\), we achieve the expected performance from \textit{HSolo}. 

\subsection{Performance on AdelaideRMF}

\begin{figure}[H]
	\centering
	\begin{subfigure}{0.4\textwidth}
		\centering
		\includegraphics[width=.8\linewidth]{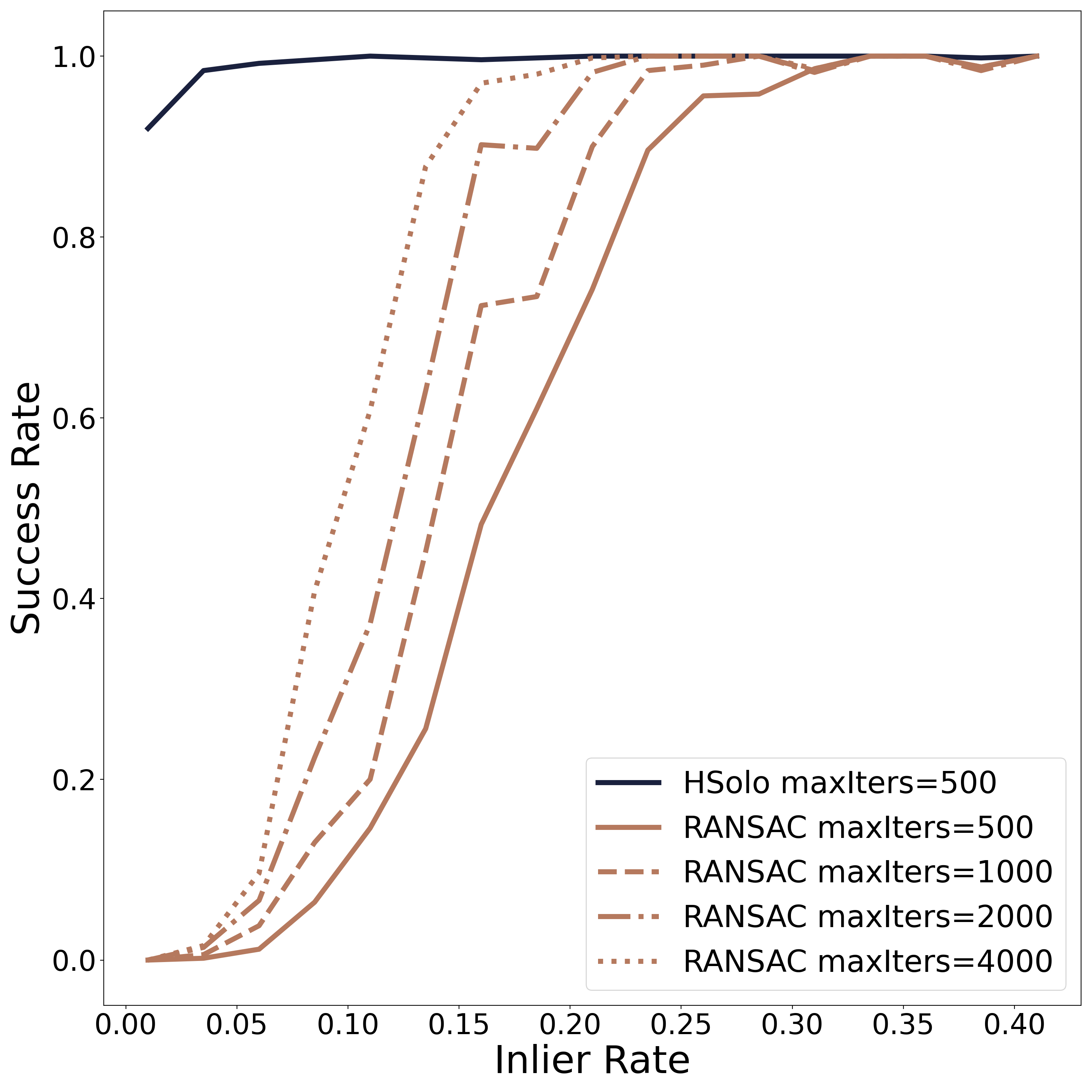}
		\caption{}
	\end{subfigure}
	\begin{subfigure}{0.4\textwidth}
		\centering
		\includegraphics[width=.8\linewidth]{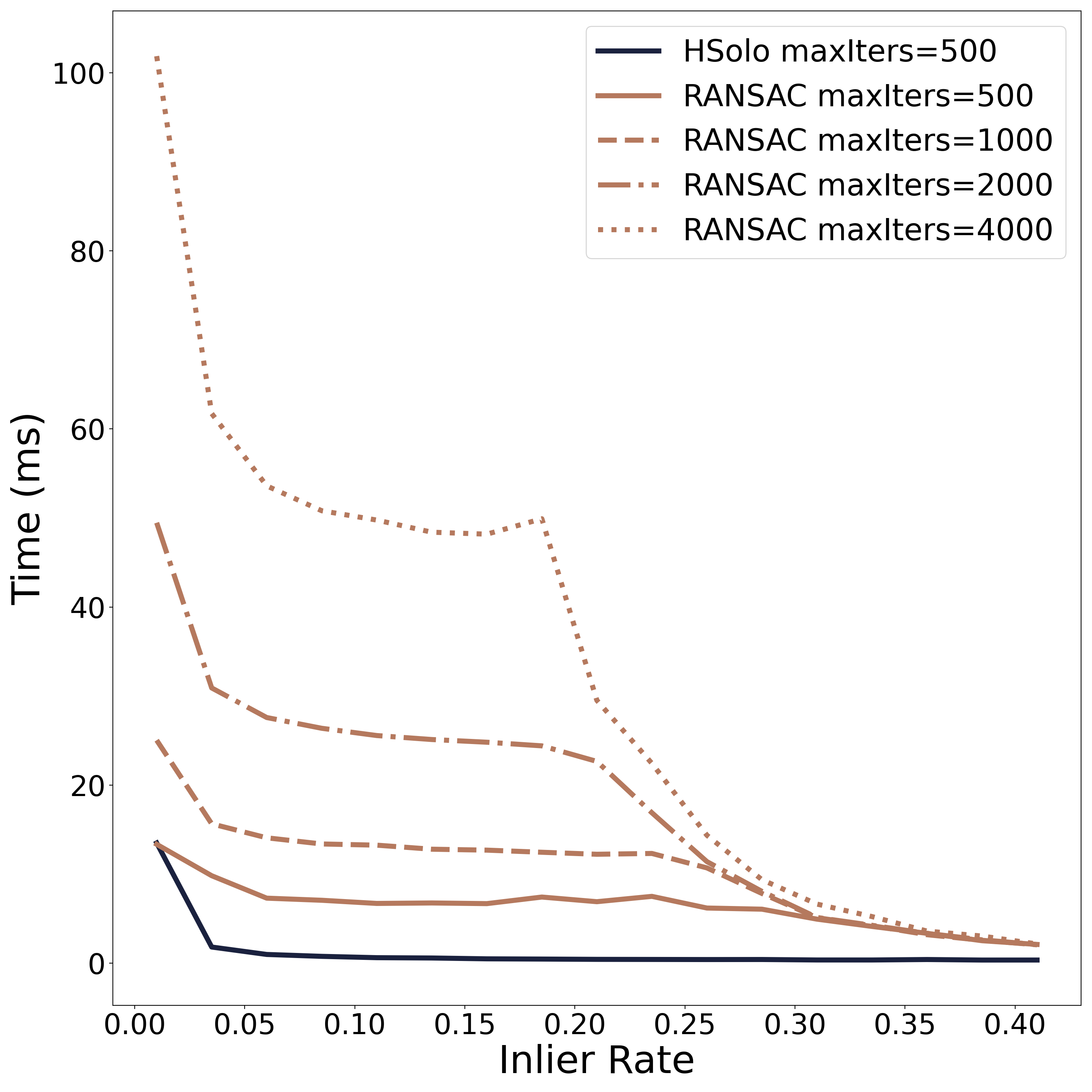}
		\caption{}
	\end{subfigure}
	\caption{(a)  Comparison of the success rate of \textit{HSolo} to \textit{OpenCV} RANSAC  for varying numbers of maximum allowed iterations for increasing values of  $w$.  (b) Comparison of the run times.  The plateau in RANSAC runtime is an artifact of the way \textit{OpenCV} dynamically recalculates its number of iterations to run. }
	\label{fig:hsolo_v_ransac}
\end{figure}

  First, we compare the performance of \textit{HSolo} to the RANSAC implementation provided by \textit{OpenCV}.  We choose a representative homography from AdelaideRMF and add or subtract random correspondences as necessary to generate inlier rates from 0.01 to 0.4.  We run 500 trials of both algorithms with the parameterization listed in Table~\ref{tab:table} and compare how often they generate a correct solution and their run times.  Both algorithms are limited to a maximum number of iterations.  Figure~\ref{fig:hsolo_v_ransac} confirms that \textit{HSolo} is able to find solutions at much lower values of $w$ while providing faster performance even with the limitation on iterations.

\begin{figure}[H]
	\centering
	\includegraphics[width=\textwidth]{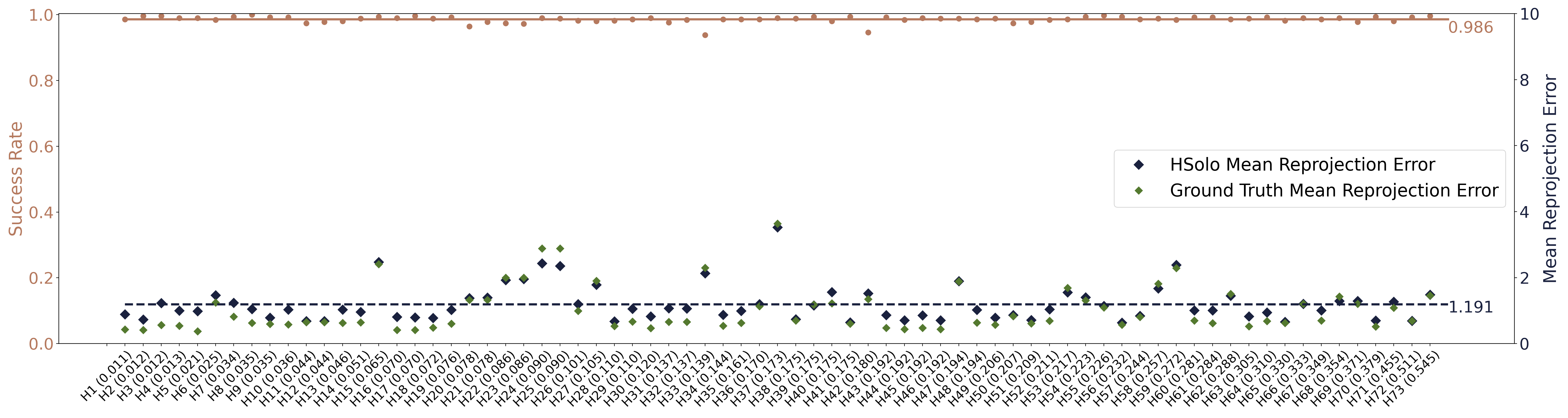}   
	\caption{Success rate (left axis) and mean reprojection error (right axis) over 500 trials for \textit{HSolo} on each homography of the AdelaideRMF data. The true inlier rate for each homography is shown in parenthesis.  The mean reprojection error using the ground truth homography is given for comparison.}\label{fig:adelaideperformance}
\end{figure}

Next, we examine the performance of \textit{HSolo} against the full AdelaideRMF dataset.  We run 500 trials for each of the homographies limiting the iterations run to the theoretical number based on \(w * 0.7\).  Due to wide variance of reprojection errors in the ground truth, it is impossible to set a global threshold on reprojection error to identify a correct result.  Instead, for each trial we calculate the mean reprojection error of the the manually defined AdelaideRMF correspondences using the resulting homography.  We apply \textit{DBSCAN}~\cite{ester1996densitybased} to break the mean errors into clusters, and assume that the largest cluster consists of correct estimates.  Thus the success rate is calculated as the size of the largest cluster compared to the number of trials.  Mean reprojection error is calculated by averaging the reprojection errors of the trials in the largest cluster.  As shown in Figure~\ref{fig:adelaideperformance}, \textit{HSolo} produces a successful result an average of \(98.6\%\) of the time.  When a correct solution is produced, the mean reprojection error is 1.191 pixels.  

\section{Conclusion}
We have presented a novel algorithm for helping to mitigate the challenges of an inlier poor correspondence set.  By leveraging the scale and rotation byproducts of affine aware feature descriptors, we are able to produce an initial estimate of the homography.  We have shown that this initial estimate is sufficient for eliminating a large percentage of the most significant outliers.  With reduced outliers, the process can be followed with standard robust estimator techniques to produce results of comparable quality.  As shown in our experiments, in such inlier poor domains, our pre-filtering based approach significantly reduces the total runtime.  Applications previously infeasible due to poor inlier rates become tractable.

\section{Acknowledgments}

We would like to thank Kurt Larson, Fred Rothganger, Stephen Rowe, Sal Sanchez, and Justin Woo for their feedback on this paper that greatly improved the exposition.  We especially would like to thank Fred for pointing out a simplification to our original method.  This report is SAND2020-9046 O.

Sandia National Laboratories is a multimission laboratory managed and operated by National Technology \(\&\) Engineering Solutions of Sandia, LLC, a wholly owned subsidiary of Honeywell International Inc., for the U.S. Department of Energy’s National Nuclear Security Administration under contract DE-NA0003525.

\bibliographystyle{unsrt}  
\bibliography{hsolo}  %%% Remove comment to use the external .bib file (using bibtex).

\end{document}